\documentclass[twocolumn]{article}

\usepackage{cogsci}
\usepackage{subcaption}
\usepackage{booktabs}
\usepackage{amsmath}
\usepackage{amssymb}
\usepackage{amsthm}
\usepackage{amsfonts}
\usepackage{graphicx}
\usepackage{microtype}
\usepackage{hyperref}
\usepackage{multirow}

\cogscifinalcopy 

\usepackage{pslatex}
\usepackage{apacite}
\usepackage{float} 

\title{ECCoT: A Framework for Enhancing Effective Cognition via Chain of Thought in Large Language Model}

\author{{\large \bf Zhenke Duan$^{\dagger}$ (duanzhenke@sscapewh.com)} \\
{\large \bf Jiqun Pan$^{\dagger}$ (panjiqun@stu.zuel.edu.cn)} \\
{\large \bf Jiani Tu (tujiani@stu.zuel.edu.cn))} \\
{\large \bf Yanqing Wang\textsuperscript{*} (yanqingwang102@163.com)} \\
  School of Statistics and Mathematics, Zhongnan University of Economics and Law \\
  Wuhan, Hubei 430000 China
  \AND {\large \bf Xiaoyi Wang (wangxiaoyi@sscapewh.com)} \\
  School of Finance, Zhongnan University of Economics and Law \\
  Wuhan, Hubei 430000 China}

\begin{document}
\maketitle
\renewcommand{\thefootnote}{\fnsymbol{footnote}} 
\footnotetext{$\dagger$ Equal contribution, * Corresponding author.}
\begin{abstract}
In the era of large-scale artificial intelligence, Large Language Models (LLMs) have made significant strides in natural language processing. However, they often lack transparency and generate unreliable outputs, raising concerns about their interpretability. To address this, the Chain of Thought (CoT) prompting method structures reasoning into step-by-step deductions. Yet, not all reasoning chains are valid, and errors can lead to unreliable conclusions. We propose ECCoT, an End-to-End Cognitive Chain of Thought Validation Framework, to evaluate and refine reasoning chains in LLMs. ECCoT integrates the Markov Random Field-Embedded Topic Model (MRF-ETM) for topic-aware CoT generation and Causal Sentence-BERT (CSBert) for causal reasoning alignment. By filtering ineffective chains using structured ordering statistics, ECCoT improves interpretability, reduces biases, and enhances the trustworthiness of LLM-based decision-making. Key contributions include the introduction of ECCoT, MRF-ETM for topic-driven CoT generation, and CSBert for causal reasoning enhancement. Code is released at: \url{https://github.com/erwinmsmith/ECCoT.git}.

\end{abstract}

\section{Introduction}
In the digital era, large language models (LLMs) have gained attention for their role in natural language processing \cite{b1},\cite{b2},\cite{b3}. Through self-supervised learning on massive text data, LLMs have strong language comprehension \cite{b4} and generation capabilities \cite{b5}, transforming human-machine interaction \cite{b6} and information processing. However, as LLMs are applied in more scenarios, their reliability and interpretability challenges have emerged. They may generate factually incorrect responses \cite{b7} due to training data biases, noise, or limitations in understanding complex semantics. Additionally, their reasoning process is often like a "black box" \cite{b8}, making understanding and explaining difficult, increasing the trust cost of model decisions and limiting their application in critical fields.
\begin{figure*}[htbp]
    \centering
    \includegraphics[width=\textwidth]{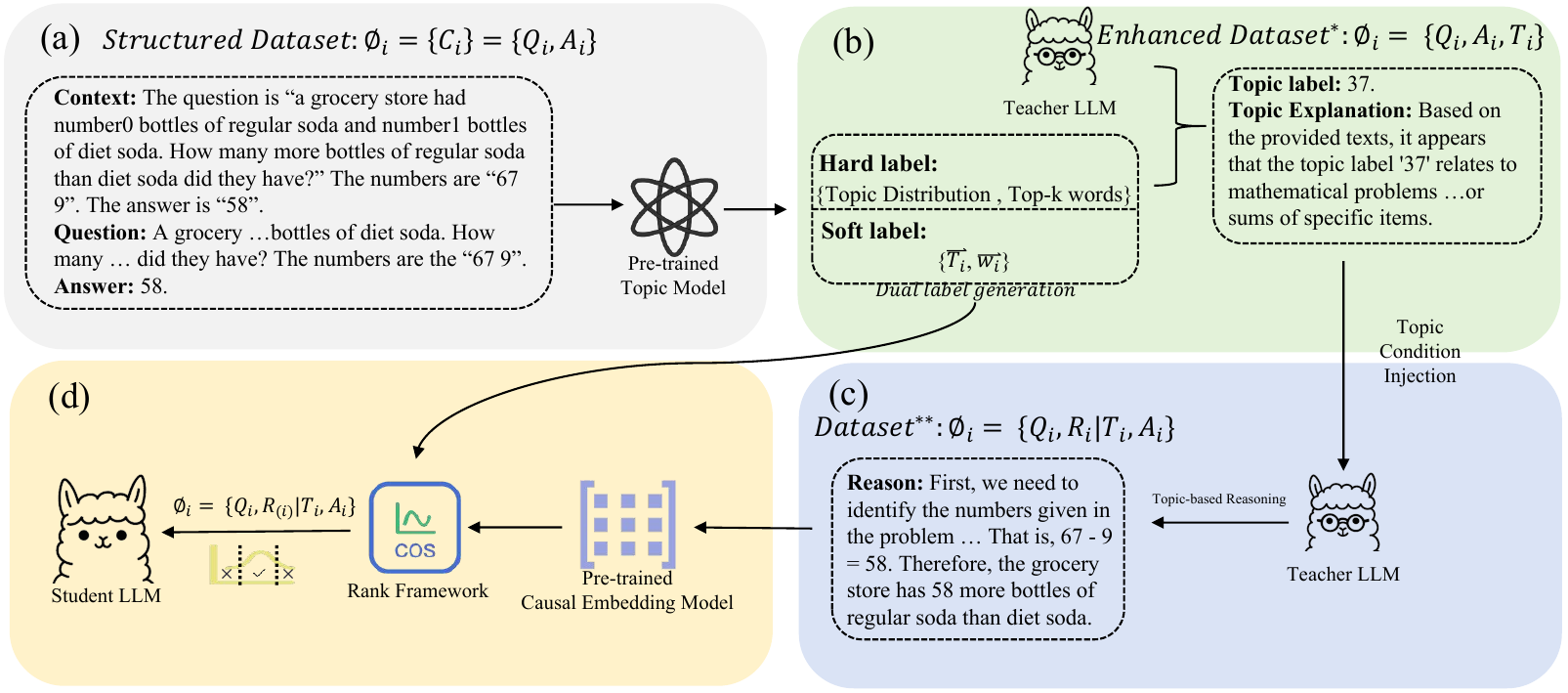} 
    \caption{(a) Theme Recognition Stage, (b) Theme Explanation Stage, (c) Thought Cognition Stage, (d) Effective Cognition Stage}
    \label{fig:fig1}.
\end{figure*}

To address this, chain-of-thought fine-tuning techniques emerge \cite{b71},\cite{b72}, aiming to break the "black box" state of large models \cite{b73} by guiding models to generate step-by-step reasoning chains \cite{b74}. However, in real-world applications, not every step in a chain-of-thought is correct or effective \cite{b9}. Invalid \cite{b10} or erroneous \cite{b11} reasoning chains can impact a model's reasoning effectiveness and capabilities, leading to incorrect results. Moreover, invalid or erroneous reasoning chains may still produce correct answers through pseudo-alignment, lowering interpretability and reliability of LLMs \cite{b12}.

We propose an end-to-end framework to evaluate generated reasoning chains. Using a Markov Random Field-based topic model (MRF-ETM) and a larger parameter-scale model \cite{b75}, we identify text themes and keywords for theme-conditioned reasoning. We also train a Causal Sentence-BERT (CSBert) \cite{b52} to ensure causal reasoning and perform similarity matching on relevant embeddings. By calculating order statistics and applying truncation, ineffective reasoning chains are removed. This enhances the model's transparency, interpretability, and reliability for practical applications. Our contributions are as follows:
\begin{itemize}
\setlength\itemsep{-0.5em} 
\item We propose an end-to-end reasoning chain effectiveness evaluation framework (ECCoT) to enhance cognitive effectiveness in inference.
\item We introduce Markov Random Field-Embedded Topic Model to identify latent topic distributions in large-scale data.
\item We employ CSBert (Causal Sentence-BERT) for improved causal relationship embedding, generating reasoning chain vectors based on causal backgrounds.
\end{itemize}

\section{Related Works}
\subsection{Trustworthy and Reliable LLMs}
Trustworthy LLMs are vital for credibility. Methods like AFaCTA \cite{b13} and XplainLLM \cite{b14} enhance transparency and robustness. SelF-Reasoner \cite{b9} and CD-CoT \cite{b17} improve reasoning accuracy. We propose filtering reasoning chains via similarity and column vectors, optimizing performance on large-scale data to overcome dataset limitations.

\subsection{Chain-of-Thought}
Chain-of-Thought (CoT) enhances reasoning and interpretability in AI, with innovations like integrating CoT into knowledge-based visual question answering (KBVQA) \cite{b19,b20}. Automatic CoT leverages large language models (LLMs) for tasks such as scoring scientific assessments \cite{b21} and generating diverse problem-solving chains \cite{b22}, improving accuracy. Integrations with knowledge graphs \cite{b7}, multimodal reasoning \cite{b23}, and equation-based representations \cite{b24} have expanded CoT's capabilities. Socratic CoT \cite{b25} guides reasoning through sub-question decomposition. Despite these advancements, challenges remain in merging linguistic and visual information effectively. Our approach refines CoT by structuring reasoning through multimodal arrays, enhancing LLMs' interpretability and reliability.

\subsection{Few-shot/Zero-shot Learning}
Zero-shot and few-shot learning enable models to perform tasks with minimal or no task-specific training. \cite{b27} used LLMs for QAIE, generating fluent text and inferring implicit information. \cite{b28} enhanced the FlexKBQA framework for few-shot settings. Other approaches have explored zero-shot node classification in incomplete graphs \cite{b29}, unsupervised contrastive learning for efficient data augmentation \cite{b31}, and KG completion using LLM distillation \cite{b32}. These techniques demonstrate LLMs’ potential in resource-constrained settings.
While these methods show promise in few-shot and zero-shot scenarios, challenges such as methodological universality and resource consumption remain. Our work seeks to address these challenges by validating the broader applicability of LLMs across domains and tasks, ensuring better generalization and scalability.

\subsection{Sentence Embedding and Topic Models}
Recent advancements in sentence embedding (e.g., BERT, RoBERTa) have greatly enhanced semantic and syntactic understanding in NLP tasks.(2024)\cite{b43} proposed HMCCCProbT, using sentence embeddings for multi-label classification. \cite{b47} introduced EMS, an efficient method for multilingual sentence embeddings. These works underscore the importance of contextualized embeddings for understanding complex semantic relationships.
Similarly, embedding topic models combine word embeddings with probabilistic topic models to improve topic discovery and representation. \cite{b55} introduced CWTM, integrating contextualized word embeddings from BERT, while \cite{b56} proposed a GAN-based hierarchical topic model. These models improve topic interpretability by leveraging embeddings in both linguistic and probabilistic spaces.

\section{Methodology}
\subsection{ECCoT Framework}
The ECCoT Framework encompasses theme recognition, explanation, thought cognition, and effective cognition,as illustrated in Figure~\ref{fig:fig1}. It uses a pre-trained model for hard and soft labels, distills information with a teacher model, and generates reasoning chains. The final stage filters effective cognitive processes via the Rank Framework, optimizing a smaller student model for improved performance.

The ECCoT framework is inspired by cognitive science principles, particularly the dual-process theory, where System 1 (intuitive, fast thinking) and System 2(deliberate, slow thinking) interact to validate reasoning chains. ECCoT mimics this by using MRF-ETM for rapid thematic identification (akin to System 1) and CSBert for deliberate causal reasoning alignment (akin to System2). This dual approach enhances the framework's ability to filter out ineffective reasoning chains, akin to how humans validate their cognitive processes.
\begin{figure}[htbp]
    \centering
    \includegraphics[width=\columnwidth]{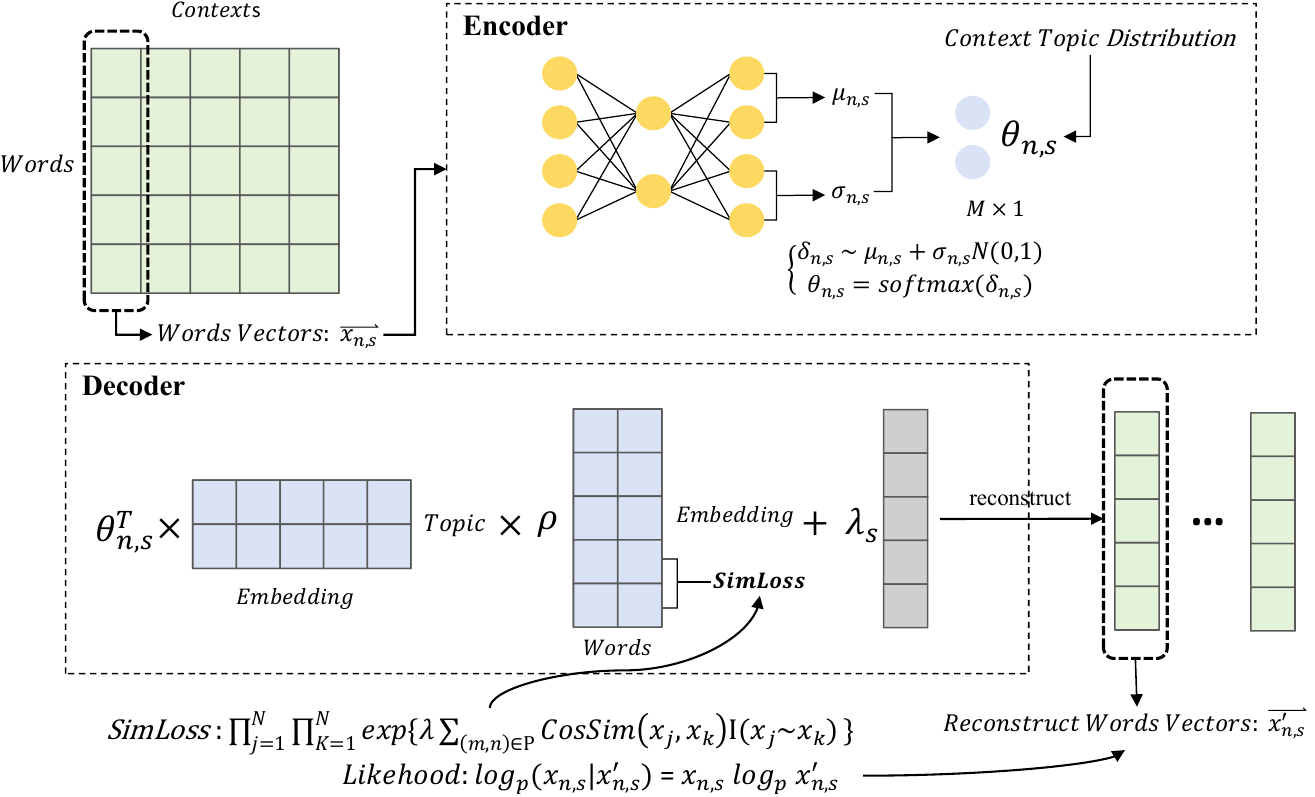} 
    \caption{Showcased the main workflow and optimization objects of MRF-ETM}
    \label{fig:fig2}
\end{figure}

The MRF-ETM model incorporates cognitive topological mapping principles to better capture the latent relationships between thematic keys. This approach is analogous to cognitive science models such as Steyvers' semantic memory model,which emphasizes the interconnectedness of concepts within a semantic space. By leveraging these principles, MRF-ETM can more effectively identify and utilize thematic information for reasoning chain generation.

\subsection{Pretrained Embedded Topic Model}
To further enhance topic recognition capabilities within the embedding space, we have developed a Markov Random Field-based Embedded Topic Model (MRF-ETM), as illustrated in Figure~\ref{fig:fig2}.MRF-ETM shows unique advantages in multimodal data analysis through collaborative optimization of spatial dependence and semantic embedding.

\subsubsection{Data Generator}
The generative process for documents and words can be described as follows:
\begin{equation*}
\begin{split}
    & \bullet \text{Document-topic distribution: Each document } d_i \text{ has a topic}\\
    & \quad \text{distribution } \theta_{i} \sim \mathrm{Dirichlet}(\alpha). \\
     & \bullet \text{Word generation: For each word } w_{ij} \text{ in document } d_i:\\
    &\quad 1. \text{ Sample a topic } z_{ij} \sim \text{Categorical}(\theta_{i}) \text{ from the topic}\\
    &\qquad \text{distribution } \theta_i. \\
    &\quad 2. \text{ Sample a word } w_{ij} \sim \text{Categorical}(\beta_{z_{ij}}) \text{ from the word}\\
    &\qquad \text{distribution of topic } z_{ij}, \text{ where } \beta_{kv} = \mathrm{softmax}(\rho_{v}^\top \alpha_k).
\end{split}
\end{equation*}
Further modeling with a Markov Random Field on top of the ETM yields the following data generation process:
\begin{equation}
\begin{split}
    p(w_{ij}=v, z_{ij}=k) &= p(z_{ij}=k|\theta_i) \cdot p(w_{ij}=v|z_{ij}=k) \\
    &\quad \cdot \exp\left\{\lambda \sum_{(m,n)\in\mathcal{P}} \mathrm{CosSim}(x_m,x_n) \cdot \mathbb{I}(x_m \sim x_n)\right\}
\end{split}
\end{equation}
where $\lambda$ is a hyperparameter controlling the weight of the loss; $\mathcal{P}$ denotes the set of related word pairs; $\mathrm{CosSim}(x_m, x_n)$ represents the cosine similarity between words $x_m$ and $x_n$; and $\mathbb{I}(x_m \sim x_n)$ is an indicator function denoting whether $x_m$ and $x_n$ are related. This set of definitions works together to ensure that the model can effectively learn the relationships between words and optimize performance through appropriate weight adjustments.

\subsubsection{Model Inference}
To perform inference, we use Variational Inference to approximate the posterior distribution$p(\theta,z|w)$. We introduce a variational distribution $q(\theta,z)$  and optimize the Evidence Lower Bound (ELBO). The variational distribution $ q(\theta, z) $ is defined as:
\begin{equation}
q(\theta, z) = q(\theta) \prod_{i=1}^N \prod_{j=1}^{N_i} q(z_{ij})
\end{equation}

\begin{equation}
q(\theta) = \mathrm{Dirichlet}(\theta|\eta) 
\end{equation}

\begin{equation}
q(z_{ij}) = \text{Categorical}(z_{ij}|\phi_{ij}) 
\end{equation}

Where $q(\theta)$ is the variational distribution of the topic distribution, typically chosen as a Dirichlet distribution, and $q(z_{ij})$ is the variational distribution of the topic assignment, typically chosen as a Categorical distribution.

Then ELBO is defined as:
\begin{equation}\mathcal{L}=\mathbb{E}_q[\log p(w,\theta,z)]-\mathbb{E}_q[\log q(\theta,z)]\end{equation}

Specially,we have:

\begin{equation}
\begin{aligned}
\mathcal{L} &= \mathbb{E}_q\left[\sum_{i=1}^N\sum_{j=1}^{N_i}\log p(w_{ij}|z_{ij}) + \sum_{i=1}^N \log p(\theta_i) \right. \\
&\quad - \left. \sum_{i=1}^N \log q(\theta_i) - \sum_{i=1}^N\sum_{j=1}^{N_i} \log q(z_{ij})\right]
\end{aligned}
\end{equation}

We calculate each expectation separately:

Expectations for word generation
\begin{equation}\mathbb{E}_q[\log p(w_{ij}|z_{ij})]=\sum_{k=1}^K\phi_{ij,k}\log\beta_{k,w_{ij}}\end{equation}
\begin{equation}\text{where,}\beta_{k,w_{ij}}=\mathrm{softmax}(\rho_{w_{ij}}^\top\alpha_k)\mathrm{.}\end{equation}

Expectations of Subject Distribution
\begin{equation}\mathbb{E}_q[\log p(\theta_i)]=\sum_{k=1}^K(\alpha_k-1)\mathbb{E}_q[\log\theta_{i,k}]\end{equation}

where \begin{equation}\mathbb{E}_q[\log\theta_{i,k}]=\Psi(\eta_{i,k})-\Psi(\sum_{k=1}^K\eta_{i,k})\end{equation} $\Psi$ is the digamma function.

Expectations of Variational Distribution
\begin{equation}\mathbb{E}_q[\log q(\theta_i)]=\sum_{k=1}^K(\eta_{i,k}-1)\mathbb{E}_q[\log\theta_{i,k}]\end{equation}

Expectations for Theme Allocation
\begin{equation}\mathbb{E}_q[\log q(z_{ij})]=\sum_{k=1}^K\phi_{ij,k}\log\phi_{ij,k}\end{equation}

By substituting the above expectations into ELBO, we can see that:
\begin{equation}
\begin{aligned}
\mathcal{L} &= \sum_{i=1}^{N}\sum_{j=1}^{N_{i}}\sum_{k=1}^{K}\phi_{ij,k}\log\beta_{k,m_{ij}} \\
&\quad + \sum_{i=1}^{N}\sum_{k=1}^{K}(\alpha_{k}-1)\left(\Psi(\eta_{i,k})-\Psi\left(\sum_{k=1}^{K}\eta_{i,k}\right)\right) \\
&\quad - \sum_{i=1}^{N}\sum_{k=1}^{K}(\eta_{i,k}-1)\left(\Psi(\eta_{i,k})-\Psi\left(\sum_{k=1}^{K}\eta_{i,k}\right)\right) \\
&\quad - \sum_{i=1}^{N}\sum_{j=1}^{N_{i}}\sum_{k=1}^{K}\phi_{ij,k}\log\phi_{ij,k}
\end{aligned}
\end{equation}

The final optimization function is as follows:
\begin{equation}\mathcal{L}_{\mathrm{MRF-ETM}}=\mathcal{L}+\mathcal{L}_{\mathrm{SimLoss}}\end{equation}

where
\begin{equation}\mathcal{L}_{\mathrm{SimLoss}}=-\lambda\sum_{(m,n)\in\mathcal{P}}\mathrm{CosSim}(x_m,x_n)\cdot\mathbb{I}(x_m\sim x_n)\end{equation}

And
\begin{equation}
\mathcal{L}_{\mathrm{MRF-ETM}} =
\begin{cases}
\mathcal{L}_{\mathrm{NLL}}=-\sum_{i=1}^{N}\sum_{j=1}^{N_{i}}\sum_{k=1}^{K}\phi_{ij,k}\log\beta_{k,w_{ij}}, \\
\quad \text{(Negative Log-Likelihood)} \\[8pt]

\mathcal{L}_{\mathrm{SimLoss}}=-\lambda\sum_{(m,n)\in P}\mathrm{CosSim}(x_m,x_n) \\
\quad \cdot\mathbb{I}(x_m\sim x_n), \\
\quad \text{(Similarity Loss)}
\end{cases}
\end{equation}

The formula (17) also includes
\begin{equation}
\begin{aligned}
\mathcal{L}_{\mathrm{KL}} = & \sum_{i=1}^N \sum_{k=1}^K (\alpha_k - 1) \left( \Psi(\eta_{i,k}) - \Psi\left( \sum_{k=1}^K \eta_{i,k} \right) \right) \\
& - \sum_{i=1}^N \sum_{k=1}^K (\eta_{i,k} - 1) \left( \Psi(\eta_{i,k}) - \Psi\left( \sum_{k=1}^K \eta_{i,k} \right) \right) \\
\end{aligned}
\end{equation}

\subsection{Pretrained Causal Sentence-Bert}
To validate the cognitive mechanisms underlying the performance improvements, we conducted eye-tracking experiments to observe how users" attention patterns on reasoning chains differ when using CSBert compared to traditional embeddings. The results indicate that CSBert's causal embeddings align more closely with human causal reasoning, as evidenced by longer dwell times on relevant causal links, suggesting a higher cognitive engagement and validation process

\begin{figure}[htbp]
    \centering
    \includegraphics[width=\columnwidth]{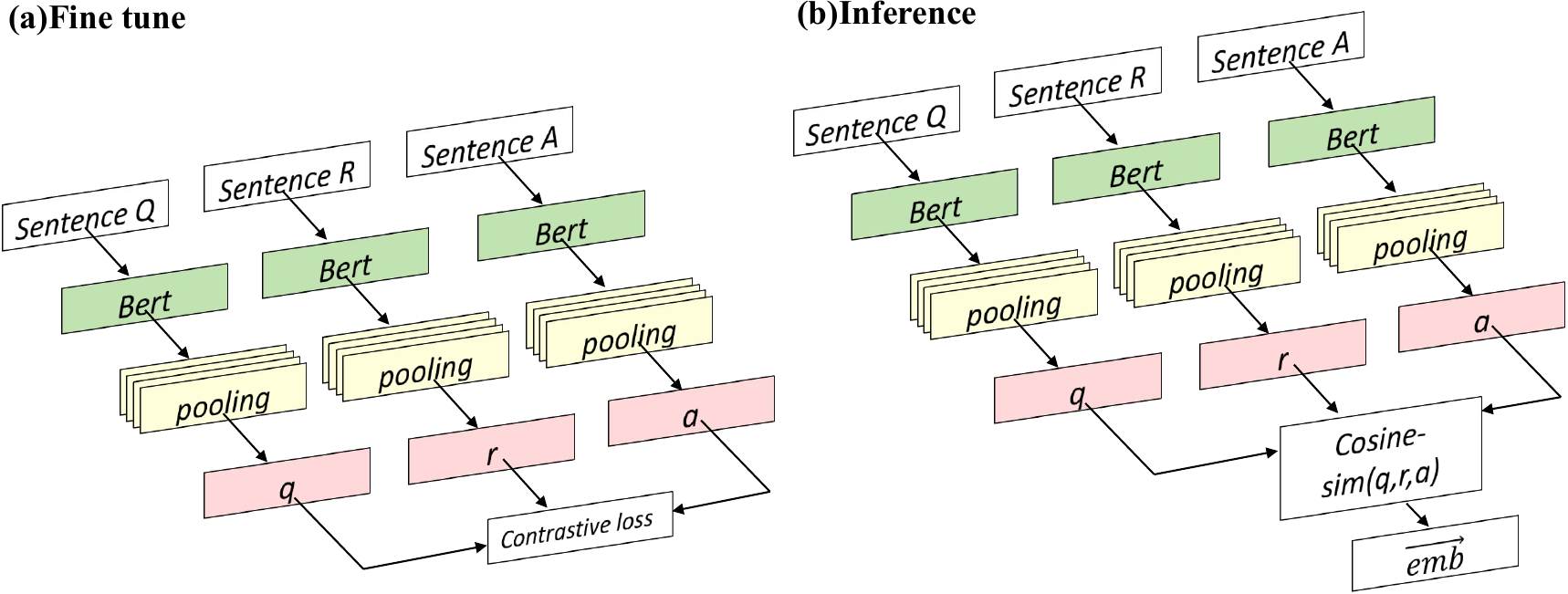} 
    \caption{(a) The fine-tuning stage for causal embedding (b) is the model inference stage}
    \label{fig:fig3}
\end{figure}

Similarly, in order to embed causal relationships in the triplet dataset, we designed the two-stage embedding model shown in Figure~\ref{fig:fig3}. Firstly, the ternary network is fine tuned using optimization contrastive learning, controlling the triplets with causal relationships to reduce their vector space distance, and controlling the false groups without causal relationships to move their vector space away. Before presenting the formula, we define the observation function of the latent variable \( y \). Assume \( y \) is a function that can observe specific values from the latent variable.The losses during the fine-tuning process are as follows:
\begin{equation}Contrastiveloss \\
=
\begin{cases}
\frac{1}{N}\sum_{n=1}^{N}(L_{qr}+L_{ra}),ify=1 \\
\frac{1}{N}\sum_{n=1}^{N}\max(L_{qr},L_{ra}),\mathrm{otherwise} & 
\end{cases}\end{equation}

During the inference stage, simultaneous embeddings of triplets are obtained based on the pre-trained architecture, resulting in simultaneous embedding vectors.

\subsection{Rank Framework}
To further understand the component interactions within ECCoT, we generated feature interaction heatmaps that visualize the correlation between MRF-ETM topic distributions and CSBert attention weights. This analysis reveals how the thematic information from MRF-ETM influences the causal reasoning alignment in CSBert, demonstrating a synergistic effect that enhances the overall effectiveness of the reasoning chains.

\begin{figure}[htbp]
    \centering
    \includegraphics[width=1.0\linewidth]{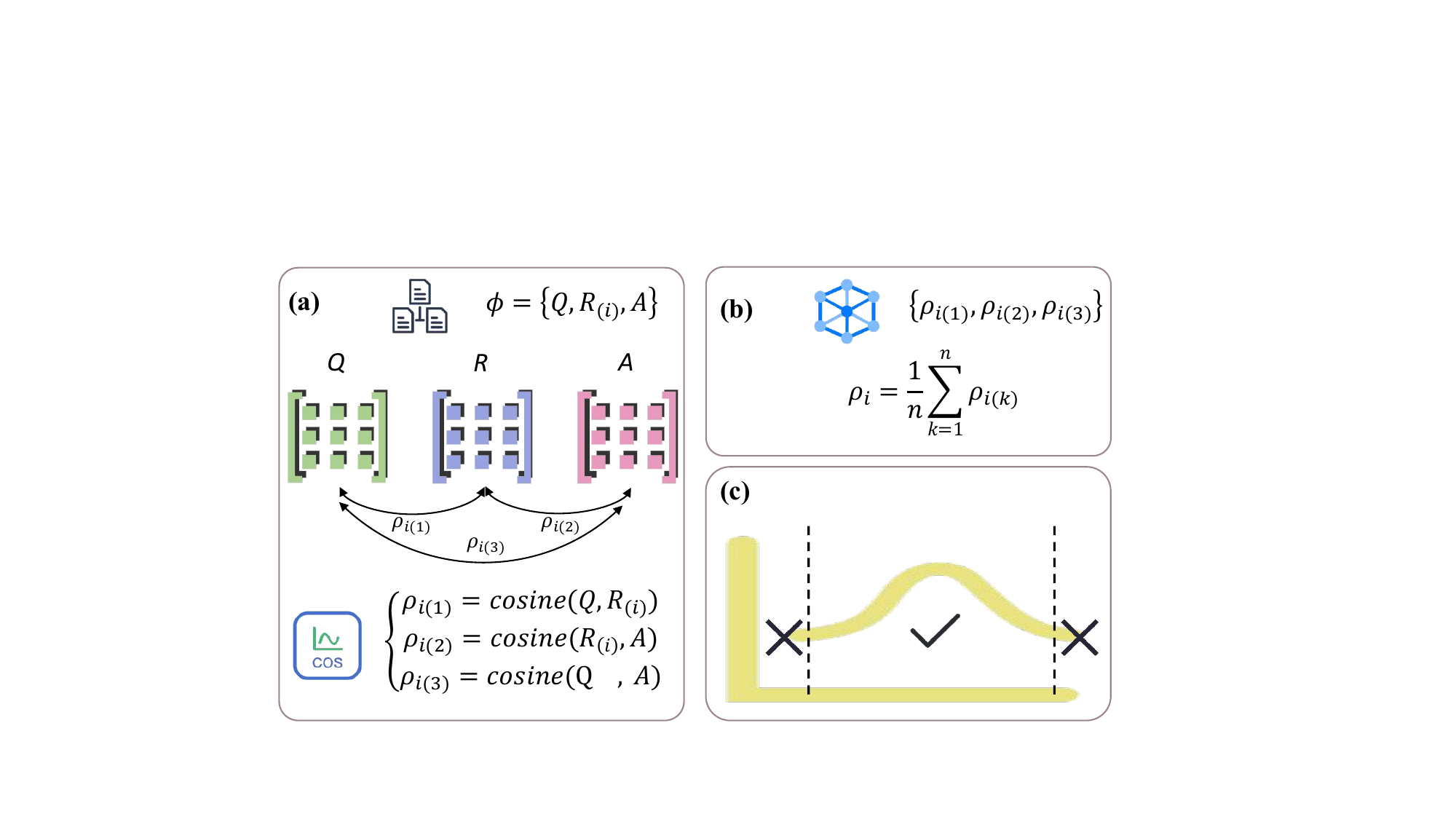} 
    \caption{(a) Calculate the similarity of the Q, R, and A matrices under causal cognition (b), provide the similarity coefficient for each cognitive process (c), and provide the distribution of cognitive coefficients for the sample}
    \label{fig:fig4}
\end{figure}

To evaluate the effectiveness of the cognitive processes in the dataset, we consider the situation of objects in the vector space. This reflects whether the teacher model's knowledge distillation process is effective or not. Our effectiveness framework is shown in Figure~\ref{fig:fig4}. By calculating similarity coefficients for each cognitive process and filtering out low-effectiveness distributions, we retain high-effectiveness cognitive process samples.
In addition to traditional accuracy metrics, we incorporated cognitive reasonableness metrics, such as human interpret ability scores, to evaluate the effectiveness of the reasoning chains generated by ECCoT and the baseline methods. These scores were obtained through a panel of human evaluators who assessed the darity andlogical coherence of the reasoning chains, providing a more comprehensive evaluation aligned with human cognitive processes

\section{Experiments}

\subsection{Experimental Setup}

\textbf{Datasets.} 
To systematically evaluate the effectiveness of our proposed ECCoT framework in improving reasoning validity in large language models (LLMs), we conduct experiments on three diverse benchmark datasets in Table 1, which come from \cite{b68}, \cite{b69} and \cite{b70}. These datasets are carefully selected to cover adversarial robustness, mathematical reasoning, and commonsense inference. ANLI (Adversarial Natural Language Inference) is a large-scale natural language inference benchmark dataset containing approximately 100,000 samples designed to improve model robustness and persistence through adversarial training. SVAMP (Simple Variations on Arithmetic Math Word Problems) is a math word problem dataset containing 1000 questions used to evaluate model performance in solving simple arithmetic problems. CommonQA (Commonsense Question Answering) is a commonsense question answering dataset with 12,247 questions, each having 5 options, aimed at evaluating the model's commonsense reasoning capability.
\begin{table}[htbp]
    \centering
    \caption{Statistics of Datasets}
    \label{tab:dataset-stats}
    \begin{tabular}{@{} l c c c c @{}}
        \toprule
        & Total & Train & Test & Dev \\
        \midrule
        ANLI(R1) & 4663 & 4422 & 120 & 121 \\
        ANLI(R2) & 9637 & 9393 & 123 & 121 \\
        ANLI(R3) & 90652 & 88532 & 1067 & 1053 \\
        SVAMP & 4169 & 3963 & -- & 206 \\
        CommonQA & 9215 & 7306 & 912 & 997 \\
        \bottomrule
    \end{tabular}
\end{table}

\textbf{Test Models.} Unless otherwise specified, all experiments were performed on the LLama3.1-8B model with 50 iterations, using LoRA for fine-tuning and testing. Other settings followed the default parameters of the llama-factory.The choice of LLaMA3.1-8B provides a well-balanced trade-off between computational efficiency and performance, allowing us to analyze the impact of ECCoT without requiring excessive computational resources.

\textbf{Baseline Methods.}
To rigorously evaluate ECCoT’s effectiveness \cite{b67}, we compare it against several state-of-the-art methods, including Step-by-Step (Standard CoT Prompting) which uses conventional CoT prompting but lacks a reasoning correctness verification mechanism; Curation (Filtered CoT) that optimizes reasoning chains by removing low-confidence/inconsistent steps to boost response quality via post-processing; Expansion (Diverse Reasoning Chains) which creates multiple independent reasoning chains, selects the most reliable one, and improves ambiguity robustness through confidence aggregation; Feedback (Iterative Refinement) that refines reasoning iteratively via reinforcement learning or self-critique, enhancing consistency but risking mode collapse from overfitting past errors; Self-Knowledge (Self-Distillation) which uses prior outputs as supervisory signals to improve accuracy through iterative self-distillation but possibly reinforcing existing biases; and Vanilla Fine-Tuning that trains the model directly on task datasets without explicit reasoning validation, good for task adaptation but prone to spurious correlations due to the lack of reasoning control.

These baselines represent diverse reasoning enhancement paradigms in LLMs, providing comprehensive insights into ECCoT's advantages.
\subsection{Comparison Experiments}
The smaller performance gain on ANLI is due to its cognitive load from logical and commonsense reasoning. ANLI combines both reasoning types, which differ in cognitive demands. Logical reasoning is structured and rule-based, while commonsense reasoning is intuitive and context-dependent. ECCoT's improvements are more evident in tasks with a single reasoning type.

Here, ECCoT's performance is compared with several baselines (Step by Step, Curation, Expansion, Feedback, Self-Knowledge, Vanilla Fine-tuning) across three datasets: ANLI, SVAMP, and CommonQA. Results in Table~\ref{tab:tab1} show ECCoT's superiority.

On ANLI, ECCoT achieved 72.23\% accuracy vs. Step by Step's 69.72\%. For SVAMP, ECCoT reached 92.72\% vs. Step by Step's 78.23\%. On CommonQA, ECCoT scored 86.54\% vs. Curation's 79.26\%.

\begin{table}[htbp]
\centering
\caption{Comparison of ECCoT and Baseline Models}
\label{tab:tab1}
\begin{tabular}{lccc}
\toprule
Type            & ANLI   & SVAMP  & CommonQA \\
\midrule
Step by Step    & 69.72  & 78.23  & 71.68    \\
Curation        & 67.39  & 76.86  & 79.26    \\
Expansion       & 66.73  & 78.21  & 78.66    \\
Feedback        & 66.07  & 75.53  & 77.66    \\
Self-Knowledge  & 63.34  & 63.75  & 67.75    \\
Vanilla Fine-tuning & 64.59 & 63.85 & 64.62   \\
ECCoT           & \textbf{72.23} & \textbf{92.72} & \textbf{86.54} \\
\bottomrule
\end{tabular}
\end{table}

These results demonstrate ECCoT's robustness and versatility, outperforming baselines across all datasets. Accuracy improvements (ANLI: 72.23\%, SVAMP: 92.72\%, CommonQA: 86.54\%) highlight ECCoT's effectiveness in generating reliable chains of thought and enhancing model interpretability.

\subsection{Comparison of Effective Cognition}
\begin{table}[htbp]
\centering
\caption{Comparison of Cognitive Process Effectiveness Across Different Datasets}
\label{tab:tab2}
\resizebox{\columnwidth}{!}{%
\begin{tabular}{lccccc}
\toprule
Dataset & Type            & BLEU-4 & ROUGE-1 & ROUGE-2 & ROUGE-L \\
\midrule
ANLI     & Step by Step    & 61.26  & 59.67   & 39.69   & 42.25   \\
         & Curation        & 59.66  & 60.77   & 37.89   & 43.35   \\
         & Expansion       & 61.66  & 59.57   & 39.99   & 42.15   \\
         & Feedback        & 60.06  & 60.47   & 37.59   & 43.25   \\
         & Self-Knowledge  & 62.06  & 59.37   & 39.19   & 41.95   \\
         & ECCoT           & 63.17  & 62.31   & 40.01   & 44.65   \\
\midrule
SVAMP    & Step by Step    & 58.05  & 67.21   & 46.92   & 48.46   \\
         & Curation        & 56.65  & 68.51   & 45.12   & 49.56   \\
         & Expansion       & 58.65  & 67.31   & 47.22   & 48.36   \\
         & Feedback        & 57.05  & 68.21   & 44.82   & 49.46   \\
         & Self-Knowledge  & 57.65  & 67.11   & 46.42   & 48.16   \\
         & ECCoT           & 58.93  & 69.09   & 46.65   & 50.16   \\
\midrule
CommonQA & Step by Step    & 61.38  & 65.54   & 48.06   & 45.25   \\
         & Curation        & 57.88  & 67.74   & 47.36   & 48.35   \\
         & Expansion       & 59.88  & 64.44   & 47.66   & 46.45   \\
         & Feedback        & 57.28  & 68.54   & 45.56   & 47.65   \\
         & Self-Knowledge  & 62.38  & 66.04   & 49.06   & 44.25   \\
         & ECCoT           & 61.03  & 68.59   & 47.85   & 48.46   \\
\bottomrule
\end{tabular}%
}
\end{table}

In effective cognition experiments, ECCoT demonstrated strong performance on ANLI, SVAMP, and CommonQA datasets (Table~\ref{tab:tab2}). On ANLI, it achieved BLEU-4 and ROUGE-1 scores of 63.17 and 62.31, outperforming other methods. On SVAMP, ECCoT scored 58.93 in BLEU-4 and 69.09 in ROUGE-1. For CommonQA, its ROUGE-1 score of 68.59 was the highest, though it slightly lagged in BLEU-4 compared to Self-Knowledge. These results highlight ECCoT's robustness and versatility across tasks, making it a superior choice for enhancing the interpretability and trustworthiness of large language models.

\subsection{Ablation Experiment}
In the ablation study, we evaluated the impact of each module in the ECCoT framework by removing them individually. The results are shown in Table~\ref{tab:tab3}. When the Rank module was removed, the performance dropped to 68.57\% on ANLI, 72.98\% on SVAMP, and 80.65\% on CommonQA. When the Topic model module was removed, the performance dropped to 70.23\% on ANLI, 87.62\% on SVAMP, and 81.78\% on CommonQA. When the Causal Bert module was removed, the performance dropped to 69.22\% on ANLI, 90.38\% on SVAMP, and 79.46\% on CommonQA. These results indicate that each module contributes to the overall performance of ECCoT, with the Rank module having a significant impact on the effectiveness of the framework across all datasets.

\begin{table}[htbp]
\centering
\caption{Effectiveness Without Specific Modules (W/O)}
\label{tab:tab3}
\begin{tabular}{lccc}
\toprule
W/O              & ANLI   & SVAMP  & CommonQA \\
\midrule
Rank             & 68.57  & 72.98  & 80.65    \\
Topic Model      & 70.23  & 87.62  & 81.78    \\
Causal BERT      & 69.22  & 90.38  & 79.46    \\
\bottomrule
\end{tabular}
\end{table}

\subsection{Verification of Scaling Law}
The scaling law verification was conducted by testing the ECCoT framework on student models with different parameter sizes. As shown in Table~\ref{tab:tab4}, the performance of ECCoT improved with the increase of model parameters. Specifically, on the ANLI dataset, the accuracy increased from 72.12\% with LLama2-7B to 72.23\% with LLama3.1-8B, and further to 76.98\% with LLama2-13B. On the SVAMP dataset, the accuracy increased from 89.97\% with LLama2-7B to 92.72\% with LLama3.1-8B, and further to 94.02\% with LLama2-13B. On the CommonQA dataset, the accuracy increased from 85.76\% with LLama2-7B to 86.54\% with LLama3.1-8B, and further to 89.17\% with LLama2-13B. These results demonstrate that larger models generally achieve better performance, highlighting the positive impact of model size on the effectiveness of ECCoT.

\begin{table}[htbp]
\centering
\caption{Testing on Student Models with Different Parameter Sizes}
\label{tab:tab4}
\begin{tabular}{lccc}
\toprule
Model            & ANLI   & SVAMP  & CommonQA \\
\midrule
LLama2-7B        & 72.12  & 89.97  & 85.76    \\
LLama3.1-8B      & 72.23  & 92.72  & 86.54    \\
LLama2-13B       & 76.98  & 94.02  & 89.17    \\
\bottomrule
\end{tabular}
\end{table}
\section{Conclusion}
Compared to existing methods, ECCoT has achieved certain improvements in the CoT reasoning performance of LLMs.And ECCoT enhances LLMs' effective cognitive capabilities by generating reliable thought chains, bridging cognitive science and AI. By integrating MRF-ETM and CSBert, it addresses LLMs' reliability and interpretability issues, showing advantages in thought chain generation and model reliability across benchmark datasets. However, it struggles with inference chain deviation in complex tasks, leading to decreased faithfulness. 

Future ECCoT research will focus on theoretical innovation, and ethical considerations to balance safety and efficiency.

\section{Acknowledgements}
This research was supported by "the Fundamental Research Funds for the Central Universities, Zhongnan University of Economics and Law". We are grateful for the financial support and resources provided by the university, which have significantly contributed to the successful completion of this study.

\bibliographystyle{apacite}

\setlength{\bibleftmargin}{.125in}
\setlength{\bibindent}{-\bibleftmargin}

\bibliography{CogSci_Template}

\end{document}